# Detect-and-describe: Joint learning framework for detection and description of objects

*Adeel* Zafar[1,*], and *Umar* Khalid[2]

[1]School of Software, Shanghai Jiao Tong University, Shanghai, China
[2]School of Electrical Engineering, Shanghai Jiao Tong University, Shanghai, China

**Abstract.** Traditional object detection answers two questions; "what" (*what the object is?*) and "where" (*where the object is?*). "what" part of the object detection can be fine grained further i-e. "what type", "what shape" and "what material" etc. This results in shifting of object detection task to object description paradigm. Describing object provides additional detail that enables us to understand the characteristics and attributes of the object ("*plastic* boat" not just boat, "*glass* bottle" not just bottle). This additional information can implicitly be used to gain insight about unseen objects (e.g. unknown object is "*metallic*", "*has wheels*"), which is not possible in traditional object detection. In this paper, we present a new approach to simultaneously detect objects and infer their attributes, we call it Detect-and-Describe (*DaD*) framework. *DaD* is a deep learning-based approach that extends object detection to object attribute prediction as well. We train our model on aPascal train set and evaluate our approach on aPascal test set. We achieve 97.0% in Area Under the Receiver Operating Characteristic Curve (AUC) for object attributes prediction on aPascal test set. We also show qualitative results for object attribute prediction on unseen objects, which demonstrate the effectiveness of our approach for describing unknown objects.

## 1 Introduction

Detecting objects in images has been one of the prime focus of researchers in the field of computer vision [1, 2, 3, 4]. Detection is the traditional naming task in which an object is identified by the name and localized by a bounding box in the image. Describing an object goes beyond the traditional naming task and identify few attributes and properties of the object [5, 6]. Detection when combined with description opens a door for whole lot of new applications in computer vision. For example, we can infer few properties and attributes of the object (e.g. "*round*" bottle), we can say what is unusual about the object (e.g. "*plastic*" car), even if an object is unrecognizable we can still say something about it (e.g. the unknown object "*has legs*", "*has head*"), and something about the object's environment can also be said (e.g. train "has leaf" means train is in trees).

In this paper, we propose an end-to-end deep learning framework for object detection and description which can detect objects and infer their attributes simultaneously, we call it

---

* Corresponding author: adeelz92@gmail.com

Detect-and-Describe (*DaD*) framework. As described in Fig. 1, image is passed through a deep convolution neural network (CNN) [7], which outputs object's bounding box and its label along with attributes of that object. Joint end-to end learning has multiple advantages over distinct learning. Firstly, simultaneous detection and attribute inference provide additional information about the identified object. This allows the user to perform appropriate actions based on the additional information gained from the object's description. For example, consider an automated driving car detects a traffic light signal, if it receives related knowledge about the state of the traffic signal ("*Red*" traffic signal), it can perform appropriate actions or take decisions accordingly (i-e "*Stop*"). Secondly, when an object is detected, a bounding box around the object is returned. Bounding box usually includes some area from the surroundings of the object as well. Simultaneously inferring attributes, along with object detection can provide some information about the surroundings as well. For example, in Fig. 1, the attribute "*has leaf*" provide information that the detected object "train" is surrounded by trees.

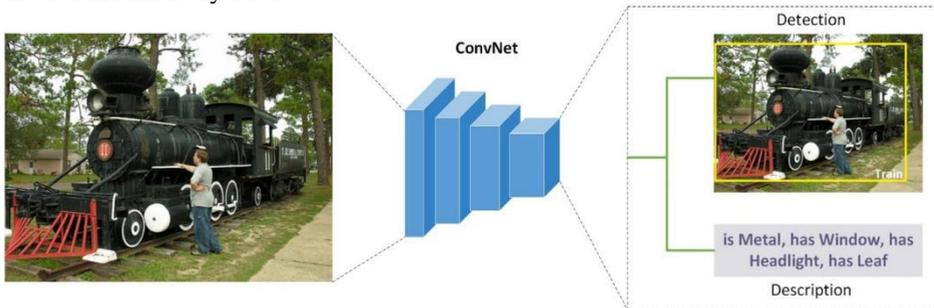

**Fig. 1.** Overview of Detect-and-Describe framework for object detection and description.

Different object categories share attributes (e.g. "Cat" and "Dog" both have "*skin*", "*leg*" and "*tail*") and same object categories may also have different attributes (e.g. "*Wooden*" boat, "*Plastic*" boat). Therefore, attribute prediction must be able to generalize accurately both within categories and across categories. In order to overcome this challenge, we learn attribute prediction separately from the object detection. It means, we learn separate convolutional features for attribute predictions and object detection. Detection uses its own sub-network to learn deep convolutional features specifically for detection task and description uses its own sub-network to learn features specifically for attributes prediction as shown in Fig. 2.

There are two types of object detectors in the deep learning literature, one-stage object detectors [4, 8, 9] and two-stage object detectors [10, 11, 12]. One-Stage object detectors simultaneously propose object proposals and then detect objects while two-stage object detectors propose object proposals in the first stage and then detect objects in the second stage. For real-time object detection, one stage object detectors demonstrated high speed but suffered from low accuracy and precision. Recently, [4] showed that one stage object detectors suffered from accuracy issues because of high density of background ("No object") proposals and low density of foreground ("object") proposals in large number of candidates bounding boxes or anchors. They introduced dynamically scaled cross-entropy loss called *Focal loss* to overcome the problem of high loss due to large density of background anchors. Since we also predict object attributes based on anchors therefore we use the *Focal loss* from [4] as the minimization objective for attributes learning. We propose Detect-and-Describe (*DaD*) framework (Fig. 2), which extends the one-stage object detector from [4] for simultaneous object detection and description.

To demonstrate the effectiveness of the proposed *DaD* framework, we perform experiments on aPascal attributes dataset [5]. We train our model on aPascal train split and report performance on aPascal test split. We achieve 97% in Area under the Receiver

Operating Characteristic Curve (AUC) for object detection on aPascal test set. We also show qualitative results of attribute predictions on unknown/unseen object categories (these categories are different from our training categories) by randomly selecting images from Google Image Search to demonstrate the generalization ability of our approach.

## 2 Related Work

Attributes detection in objects goes almost as back as object detection in the history of computer vision. As attributes are merged within the object itself, researchers have been learning the role of attributes detection in object detection e.g. [13]. In feature integration theory of attention [14], Treisman and Gelade theorized that human brain recognizes objects in two stages. At first stage, human brain perceives information about the basic features of the visual area in focus and then in second stage, focused visual attention is used to combine the individual features of an object to recognize the object. It states that object attributes play an important role in understanding the nature of the object. Since the inception of feature integration theory of attention, researchers in computer vision have adopted attributes recognition for object categorization as well as object description.

Ferrari and Zisserman [15] presented a probabilistic generative model to learn simple color and pattern attributes in image segments in a weakly supervised setting using Google image search. Later, [5] presented a new dataset for attributes based on Pascal VOC 2008 [16] object detection dataset and presented a classification model that learned to predict attributes using HOG features [17]. They also learned to categorize objects based on the annotated semantic features and learned discriminative features. The most relevant work is [18] which learned to predict object location via iterative procedure in a weakly supervised manner, where images are only labelled with categories but not object locations. They use initial model to predict saliency score for the object and use visual cues from attributes model to improve the model for object detection (*location and class prediction*), which is their prime goal. In contrast to their work, our goal is to improve the performance of attributes prediction while retaining object detection capability using end-to-end deep learning model. In *DaD* framework, we build object and attributes subnetworks on top of feature pyramid network [12] to detect objects and their attributes in the anchor-boxes/bounding-boxes.

## 3 Detect-and-Describe Framework

In one stage object detectors, multiple anchor boxes of different sizes and aspect ratios are proposed on each spatial location of some deep CNN layer of the backbone network. Anchors are assigned positive (*"Object"*) or negative (*"No object"* or *"Background"*) labels based on IoU value with the ground truth bounding box. Final anchors for the output are chosen based on top objectness score and filtering best anchors based on non-maximum suppression. The class label and bounding box offsets are computed for these anchors. In *DaD* framework, in addition to object's label and bounding box offsets, the log probability for multi-label attributes is also computed for final anchors. The attribute set for the detected object is the attributes probability score for each final anchor box, in which score of 0 depicts *attribute not present* and score of 1 depicts *attribute is present*. As most of the anchors are assigned negative label due to low IoU with the ground truth bounding box, they produce a negative learning signal which results in high loss while training. [4] presented focal loss to overcome this problem.

### 3.1 Focal Loss

The standard cross entropy ($CE$) loss for binary classification is defined as

$$CE(p,y) = \begin{cases} -\log(p) & if\ y = 1 \\ -\log(1-p) & otherwise \end{cases} \quad (1)$$

where $y \in \{\pm 1\}$ specifies ground truth label and $p \in [0, 1]$ is the model's estimated probability for the class with label $y \in 1$.

The $CE(p,y)$ can be rewritten as $CE(p_t) = -\log(p_t)$ where $p_t$ is defined as:

$$p_t = \begin{cases} p & if\ y = 1 \\ 1-p & otherwise \end{cases} \quad (2)$$

Then the *focal loss* is defined as

$$FL(p_t) = -\alpha(1-p_t)^\gamma \log(p_t) \quad (3)$$

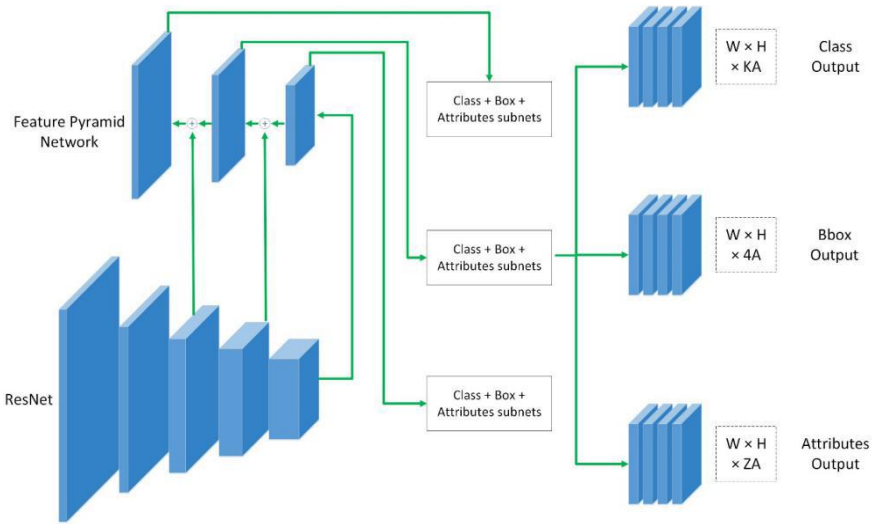

**Fig. 2.** In *DaD* framework, image is first passed through Resnet [19] which consists of 5 convolution blocks, a feature pyramid network [12] is built on top of the final convolution layers of each block (Conv3, Conv4 and Conv5 layers) of ResNet. Each layer in feature pyramid is connected to their respective bounding box, class, and attributes subnets, where $W = width$, $H = height$ of the last convolution layer of subnets and $K =$ number of classes, $4 =$ Bbox offsets, $A =$ number of anchors and $Z =$ number of attributes.

where $\alpha \in [0, 1]$ is the weighting factor to address class imbalance (high ratio of negative anchors w.r.t positive anchors) and $\gamma \geq 0$ is the tuneable focusing parameter that can decrease or increase the weight of easily classifiable examples. As large number of anchors in one stage detectors are backgrounds, they are easy to classify, so when background anchor is correctly classified then $p_t \to 1$ and $(1 - p_t)$ is small which down-weights the loss for background classes. This way negative signal due to background anchors is eliminated.

**3.2 Model Architecture**

We extend one stage object detector called RetinaNet detector from [4] for attributes recognition. RetinaNet consists of backbone network and two task specific sub-networks.

Backbone network is the convolutional neural network (ResNet [19] in our case) and two subnetworks are class label predictor and bounding box predictor. A feature pyramid network (FPN) [12] is built on top of the backbone network and these sub-networks are connected to each layer of the FPN. In addition to the class sub network and bounding box sub-network, we add an attribute sub-network, as shown in Fig. 2.

### 3.2.1 Attribute subnetwork

The design of the attribute sub-network is exactly the same as class subnet and box subnet. A fully convolutional network (FCN) is built on top of each layer of FPN, it takes $C$ input channels from the given layer of the pyramid and convolve the input four times using $3 \times 3$ kernel and $C = 256$ channels each time. Each convolution layer in this FCN is followed by ReLU activations. The sigmoided final layer outputs $ZA$ channels expressing probabilities of $Z$ attributes for $A$ anchors on each spatial location of the input. The translation-invariant anchors of multiple sizes ($32^2$ to $512^2$) and aspect ratios (1:2, 1:1, 2:1) are used to detect objects of different sizes and shapes. Our attribute predictor subnetwork uses the same anchors to predict attributes for the object.

## 4 Experiments

In this section, we report the details of several experiments to evaluate the performance of *DaD* framework on the attribute prediction task using the aPascal dataset [5]. We first perform training on aPascal train split and choose the optimal parameters. In the end, we perform testing on aPascal test to report within category generalization and randomly chosen images from Google to report across category generalization.

### 4.1 Dataset

The aPascal dataset [5] is built on top of Pascal VOC 2008 object classification and detection dataset [16]. The dataset consists of 2113 images in train split with 6340 object instances and 2227 images in test split with 6355 object instances. aPascal dataset is annotated with 64 different types of sematic attributes for each object instance in both splits. These attributes contain shape, parts and material attributes. We choose images which are closely associated to aPascal classes but different object categories from Google to report across category generalization. We choose images of donkey, monkey, goat, wolf, jetski, zebra, centaur, mug, statue, building, bag and carriage. The classes in test images are closely related to classes in aPascal, e.g. "wolf" in test images is related to "dog", "statue" is related to "person" etc in aPascal, therefore attributes should generalize well to test images as well.

### 4.2 Experimental setup

We use aPascal train split for training and use 25% randomly chosen subset of aPascal test for validation. We use Adam optimizer [20] with $\beta 1 = 0.9$ and $\beta 2 = 0.99$. The base learning rate is set to $1 \times 10^{-5}$ which decays by a factor of $0.5$ if training loss is not reduced after 2 epochs. We use ResNet-152 as backbone network and use batch size of 1 image because of high memory consumption of the model. Image is resized to minimum side of 800 and maximum side of 1333 to facilitate detection of small objects. We terminate training at 50 epochs and choose the best model for testing based on validation accuracy. We report results on aPascal test. All experiments are implemented in Keras [21].

**Table 1.** AUC score for different types of backbone networks in feature pyramid network.

| Backbone Network | AUC |
|---|---|
| ResNet-50 | 92.71% |
| ResNet-101 | 94.64% |
| VGG-Net | 93.11% |
| ResNet-152 | **97%** |

## 5 Results

### 5.1 Performance on aPascal

As mentioned earlier, within category generalization can be checked on dataset where distribution of instances is from the same set of classes for train and test splits. Therefore, to test within category generalization, we train on aPascal train split and report performance on aPascal test split. We use area under ROC curve (AUC) as performance metric for attribute predictions because it is class invariant. We achieve 97% in AUC for attributes prediction. Table 1 shows different ablations of experiments on different backbone networks.

### 5.2 Detecting and describing objects

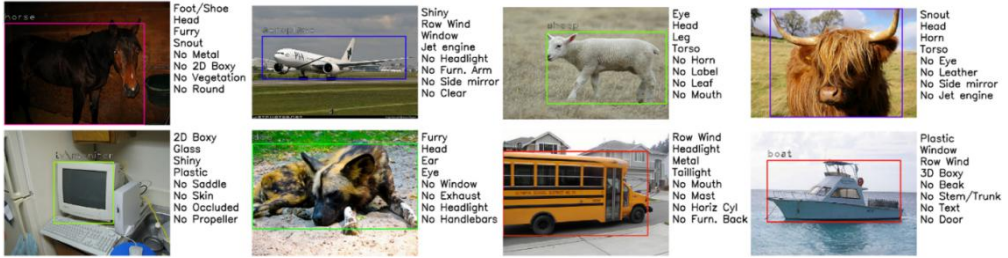

**Fig. 3(a).** This figure shows detection and attributes for randomly chosen images for some of the classes in aPascal. 8 randomly chosen attributes for each of the image are shown. Attributes starting from *"No"* means attribute is not present, otherwise attribute is present.

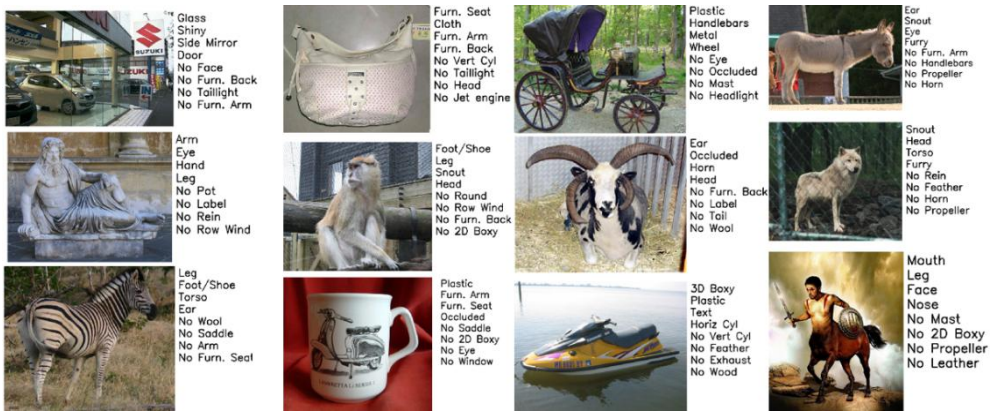

**Fig. 3(b).** This figure shows attribute predictions on images of unknown objects.

We now show some qualitative results on real images from aPascal test to verify the detection and description ability of *DaD* framework in Fig. 3(a). Some interesting and unusual results can be seen in the images, e.g. in the "cow" image (1st row, 4th column), the attribute *"No Eye"* is predicted, because eyes are covered by hair. In the same image, attribute *"No Leather"* is predicted, because this cow's skin is hairy and leather surface is invisible. Similarly, in the "sheep" image (1st row, 3rd column), *"No Mouth"* is predicted because sheep's mouth is too small and almost invisible. This shows the capability of *DaD* framework to recognize what is unusual about the object while retaining the detection ability.

### 5.3 Describing unknown objects

We show 12 different images for each of the test classes mentioned in Section 4.1 in Fig. 3(b). Model is able to generalize attributes learned from other categories to new categories fairly well. For example, in "goat" image (2nd row, 3rd column) model predicts *"Horn"* attribute for goat, which during training is learned from "cow" category. Similarly, in "carriage" image (1st row, 3rd column), model is able to predict the presence of *"wheel"*, although these wheels are wooden, while during training is learned from "cars" which have metal and rubber wheels. New classes which do not have any close association with the training classes, do not generalize well, e.g. in "building" image (1st row, 1st column), system predicts attributes for the known objects, but these known objects are in vicinity of the building therefore it predicts some attributes for the building such as *"Glass"*.

## 6 Conclusion

In this paper, we present Detect-and-Describe (*DaD*) framework to simultaneously detect objects and recognize their attributes. Compared with existing approaches to predict attributes, the *DaD* framework achieves significant performance improvement both within object categories and across categories. The framework can recognize attributes of the unknown objects fairly well due to high generalization ability of our approach. *DaD* framework is an extension of one stage object detector called "RetinaNet", in which we add an additional attribute sub-network to predict attributes, along with class label and bounding box. We find out that attributes learning can be transferred fairly well to new objects without having prior knowledge about them, but generalization is better for those objects which have same symmetry, material, body parts and texture to the learned objects which is an understandable limitation.